\documentclass[letterpaper, 10 pt, conference]{ieeeconf}  

\IEEEoverridecommandlockouts                              

\usepackage{amsmath}
\usepackage{amssymb}
\usepackage{svg}
\usepackage{graphbox}
\usepackage[normalem]{ulem}
\usepackage[hidelinks]{hyperref}

\newcommand{\qvec}[1]{\mathbf{#1}}
\newcommand{\qmat}[1]{\mathbf{#1}}
\newcommand{\idx}[1]{_{\mathrm{#1}}}

\newcommand{\qRealNumbers}{\mathbb{R}}
\newcommand{\qPositiveNaturalNumbers}{\mathbb{N}_{\geq 0}}
\newcommand{\qNaturalNumbers}{\mathbb{N}}

\newcommand{\qyest}{\hat{\qy}}
\newcommand{\qztraining}{\bar{\qz}}
\newcommand{\qVtraining}{\bar{\qV}}
\newcommand{\qpredictedmean}{\boldsymbol{\mu}}
\newcommand{\qpredictedvar}{\boldsymbol{\Sigma}}
\newcommand{\qgrad}{\boldsymbol{\nabla}}
\newcommand{\qdeltau}{\boldsymbol{\Delta}\qu}

\newcommand{\qreperror}{e_\mathrm{R}}
\newcommand{\qy}{\qvec{y}}
\newcommand{\qu}{\qvec{u}}
\newcommand{\qp}{\qvec{p}}
\newcommand{\qx}{\qvec{x}}
\newcommand{\qf}{\qvec{f}}
\newcommand{\qr}{\qvec{r}}
\newcommand{\qe}{\qvec{e}}
\newcommand{\qm}{\qvec{m}}
\newcommand{\qv}{\qvec{v}}
\newcommand{\qz}{\qvec{z}}

\newcommand{\qC}{\qmat{C}}
\newcommand{\qV}{\qmat{V}}
\newcommand{\qK}{\qmat{K}}
\newcommand{\qI}{\qmat{I}}
\newcommand{\qL}{\qmat{L}}
\newcommand{\qP}{\qmat{P}}
\newcommand{\qS}{\qmat{S}}
\newcommand{\qW}{\qmat{W}}

\newcommand{\qTwoNorm}[1]{\left\| {#1} \right\|_{2}}

\title{\LARGE \bf
AI-MOLE: Autonomous Iterative Motion Learning for Unknown Nonlinear Dynamics with Extensive Experimental Validation*
}

\author{Michael Meindl$^{1}$, Simon Bachhuber$^{2}$, and Thomas Seel$^{1}$
\thanks{*This is a preprint of the final manuscript published at \url{https://doi.org/10.1016/j.conengprac.2024.105879}}
\thanks{$^{1}$Michael Meindl and Thomas Seel are with the Institute of Mechatronic Systems, Leibniz Universit\"at Hannover, 30823 Garbsen, Germany, \{michael.meindl, thomas.seel\}@imes.uni-hannover.de}
\thanks{$^{2}$Simon Bachhuber is with the Department Artificial Intelligence in Biomedical Engineering, FAU Erlangen-N\"urnberg, 91052 Erlangen, Germany}}

\begin{document}

\maketitle
\thispagestyle{empty}
\pagestyle{empty}

\begin{abstract}
This work proposes \emph{Autonomous Iterative Motion Learning} (AI-MOLE), a method that enables systems with unknown, nonlinear dynamics to autonomously learn to solve reference tracking tasks.
The method iteratively applies an input trajectory to the unknown dynamics, trains a Gaussian process model based on the experimental data, and utilizes the model to update the input trajectory until desired tracking performance is achieved.
Unlike existing approaches, the proposed method determines necessary parameters automatically, i.e., AI-MOLE works plug-and-play and without manual parameter tuning.
Furthermore, AI-MOLE only requires input/output information, but can also exploit available state information to accelerate learning.

While other approaches are typically only validated in simulation or on a single real-world testbed using manually tuned parameters, we present the unprecedented result of validating the proposed method on three different real-world robots and a total of nine different reference tracking tasks without requiring any a priori model information or manual parameter tuning.
Over all systems and tasks, AI-MOLE rapidly learns to track the references without requiring any manual parameter tuning at all, even if only input/output information is available.
\end{abstract}

\section{Introduction}
\begin{figure*}[h!]
\centering
\includegraphics[width=0.95\linewidth]{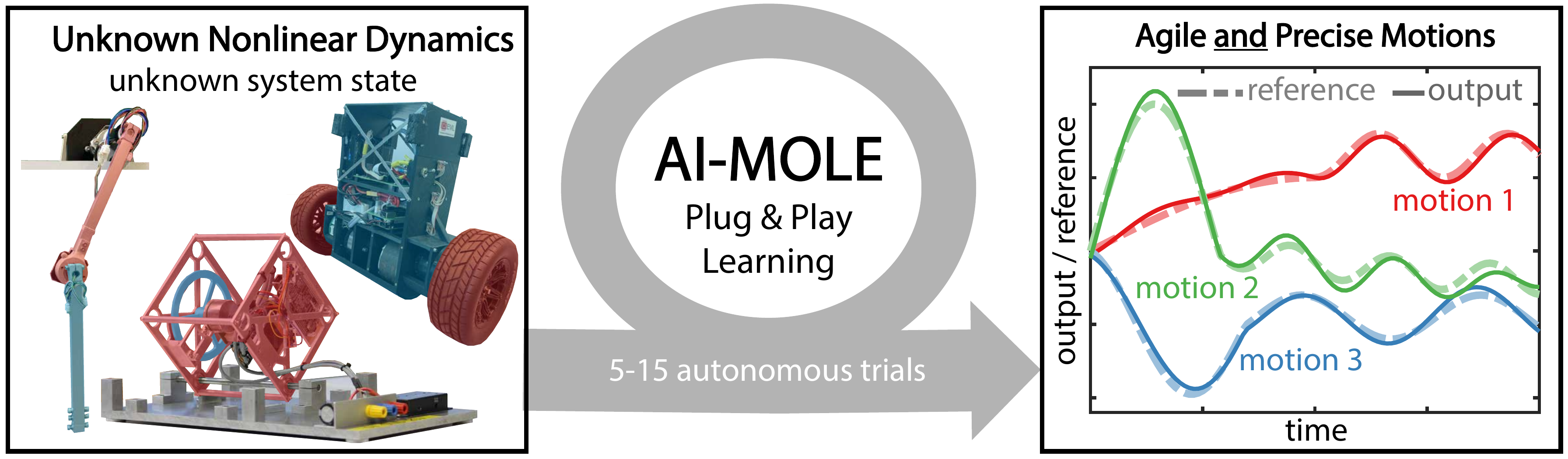}
\caption{The method proposed in this work, called \emph{AI-MOLE}, enables robotic systems with unknown, nonlinear dynamics to autonomously learn to track reference trajectories that lead to agile and dynamic motions. Learning only requires 5--15 trials, works in real-world environments and without any manual parameter tuning. }
\label{fig:fig1}
\end{figure*}

Autonomous robots are enriching a variety of application domains with examples ranging from industrial production systems \cite{MICHALOS2016195}, across rescue missions \cite{Murphy2004}, up to exoskeletons in biomedical engineering \cite{5975442}.
All of these use cases have in common that complex tasks can be solved by combining careful planning with the precise execution of motions. 
The classic approach to perform these motions consists of modeling the system, designing a controller, and tuning the necessary parameters in experiment, as, e.g., in model predictive control \cite{Apgar2018}, adaptive control \cite{Dong2005}, or robust control \cite{Golovin2019}.
While this approach enables robots to perform challenging motions and to, subsequently, solve complex tasks, it also comes at the cost of requiring enormous expert time and knowledge in order to design the controllers. 

To mitigate this downside, a variety of works has been devoted to developing learning methods that not only enable robots to autonomously perform but to autonomously \emph{learn to perform} challenging motions.
Approaches such as iterative learning control \cite{bristow2006survey,ahn2007iterative,xu2011survey}, data-driven model-predictive control \cite{berbkohlal, rosolia2018data,prag2022toward} or reinforcement learning \cite{zhao2020sim, kober2013reinforcement, brunke2022safe} have obtained impressive achievements such as guaranteeing stability or solving complex motion tasks such as walking.
However, these achievements typically come at the cost of restrictive model assumptions, requiring system- or task-specific knowledge or being limited to simulated environments.
But in order to enable robots to learn motions in a truly autonomous fashion, methods have to ultimately possess two major properties.
First, learning methods have to be real-world applicable, i.e., the methods have to be sufficiently fast and robust to solve learning problems not only in simulated but also in real-world environments.
Second, learning methods have to be plug-and-play applicable, i.e., the methods must not require manual tuning of parameters, a priori model information, nor system- or task-specific knowledge in order to be applicable to a novel system or task.

The latter of these two properties already poses a strong requirement w.r.t. the state of the art, because most methods of ILC \cite{angelini2018decentralized, sferrazza2017trajectory} and MPC \cite{9361343, carron2019data} that have been validated by real-world experiments typically require a priori model information and/or manual parameter tuning.
Real-world applicability also poses a strong requirement w.r.t. the state of the art, because methods of RL \cite{deepmind_suite, heess2017emergence, 9028188}, data-driven ILC \cite{huo2019model, 8333746, 8745534} and data-driven MPC \cite{9303965, 9756053} are predominantly validated by simulation and \emph{not} real-world experiments.
Nonetheless, some methods of RL have been validated in real-world experiments \cite{PETERS2008682,NIPS2008_7647966b,kormushev2013reinforcement}, but these methods require system- and task-specific prior knowledge in the form of good initial policies and are, hence, not plug-and-play applicable.
One example is given by \cite{smith2022walk}, in which a quadruped learns to walk in 20 minutes but only by exploiting system- and task-specific prior knowledge.
Similarly, there are data-driven ILC  \cite{8902224, 9063662, 6148318, Huo2020} and data-driven MPC methods \cite{berbkohlal, 8691673} that have at least been validated on a single experimental task, but also require manually tuned learning parameters, i.e., the methods are not plug-and-play applicable.
Lastly, there are hybrid learning methods that combine concepts of machine learning and systems and control.
However, these methods either have been validated only on a single experimental system and task \cite{9361343, berkenkamp2015safe}, or manual parameter tuning was required to apply them to multiple experimental test-beds \cite{deisenroth2011pilco, HESSE201815}.
To the best of the authors' knowledge, there exists only one RL method \cite{wu2023daydreamer} that works in a truly plug-and-play fashion and has been validated on multiple real-world systems. 
However, the method still required 1--10 hours of system interaction to learn the solution to a task.

In summary, data-driven methods have aimed to enable learning in \emph{single} tasks and \emph{single} applications under restrictive assumptions such as requiring a priori model information, manual parameter tuning, or system and task-specific knowledge.
Moreover, there is a serious lack of validation w.r.t. real-world and plug-and-play applicability.
In this manuscript, we build on our previous work \cite{10.3389/frobt.2022.793512} and propose a novel GP-based method for rapid, plug-and-play motion learning, called \emph{Autonomous Iterative Motion Learning} (AI-MOLE), see Figure \ref{fig:fig1}.
AI-MOLE is autonomous in the sense that it self-reliantly determines necessary parameters and, hence, enables plug-and-play motion learning.
Furthermore, AI-MOLE only requires input/output information, i.e., it can also be applied to systems such as soft robots where no state vector is known or measured. 
Nonetheless, the method is also capable of exploiting state knowledge to accelerate the speed of learning if a state vector is known and measured. 
In this work, we present the unprecedented result of applying the proposed learning method to three different real-world systems and solving three reference-tracking tasks per real-world system in a truly autonomous fashion.
We further show that the method only requires input/output information to solve the nine reference tracking tasks, but that the method is also capable of accelerating the speed of learning by exploiting state information if available.

\section{Problem Formulation}
Consider a system that can autonomously attempt a reference tracking task, e.g., a robot trying to perform a desired maneuver.
We assume that the system's output, e.g., a joint angle or position, can be influenced by an input signal, e.g., a  motor torque, and that the relation of these variables is deterministic, causal, and time-invariant.
However, we \emph{do not} assume that a model of the nonlinear dynamics is available. 

Formally, consider a discrete-time, single-input, single-output, repetitive system with a finite trial duration of $N\in\qPositiveNaturalNumbers$ samples, and, on trial $j\in\qPositiveNaturalNumbers$ and sample $n\in[1,N]$, output variable $y_j(n)\in\qRealNumbers$, respectively input variable $u_j(n)\in\qRealNumbers$.
The samples are collected in the so called output trajectory $\qy_j\in\qRealNumbers^N$, respectively input trajectory $\qu_j\in\qRealNumbers^N$, i.e., 
\begin{align}
    \qy_j &:= \begin{bmatrix}
    y_j(1) & y_j(2) & \dots & y_j(N)
    \end{bmatrix}^\top \,,
    \\ 
    \qu_j &:= \begin{bmatrix}
    u_j(1) & u_j(2) & \dots & u_j(N)
    \end{bmatrix}\,.
\end{align}
With respect to the system dynamics, two different cases are considered. 
First, there is the case in which only the output variable $y_j(n)$ is known and measured, and the dynamics can, without loss of generality, be written in the lifted form 
\begin{equation}\label{eq:lifted_dynamics}
\forall  j\in\qPositiveNaturalNumbers,\quad \qy_j = \qp(\qu_j)\,,
\end{equation}
where $\qp: \qRealNumbers^N\mapsto \qRealNumbers^N$ is the unknown, trial-invariant, nonlinear lifted dynamics.
Second, there is the case in which the system's state vector $\qx_j(n)\in\qRealNumbers^M$ is known and measured, and the state dynamics are given by, $\forall j\in\qPositiveNaturalNumbers$,
\begin{align}\label{eq:statespace_dynamics}
    \qx_j(n+1) &:= \qf(\qx_j(n), u_j(n))
    \\
    \label{eq:output_equation}
    y_j(n) &:= \qC\qx_j(n)\,,
\end{align}
where $\qf:\qRealNumbers^M\mapsto \qRealNumbers^M$ is the unknown, trial-invariant, nonlinear state dynamics and $\qC\in\qRealNumbers^{1\times M}$ is the known output matrix, i.e., the output variable is a linear combination of the state variables.
Note that in both scenarios, the dynamics can also - without loss of generality - contain an underlying feedback controller.

We assume that the desired motion is defined by a reference trajectory $\qr\in\qRealNumbers^N$, and the control task consists in making the output trajectory $\qy_j$ precisely track the reference trajectory $\qr$. 
The learning task consists in updating the input trajectory $\qu_j$ from trial to trial such that the output trajectory $\qy_j$ converges to the desired reference trajectory $\qr$.
Tracking performance is measured by the error trajectory
\begin{equation}
    \forall j\in\qPositiveNaturalNumbers,\quad \qe_j:= \qr-\qy_j\,.
\end{equation}

The problem considered in this work consists in developing a learning method that updates the input trajectory on each trial such that the error norm decreases.
Learning performance is evaluated based on the progression of the error norm across trials.
Ideally, the error norm should decline quickly and monotonically. 
The learning method must not require a priori model information of the plant dynamics.
To support plug-and-play application, the learning method must either provide robust default parameters or self-tune, and \emph{must not} require manual parameter tuning.
The method must be applicable to systems, for which only the output variable is measured, but the framework must also be capable of levering state information if available.

\section{Proposed Learning Method}
\begin{figure*}
    \centering
    \includegraphics[width=\linewidth, trim={0cm 0cm 0cm 5cm}, clip]{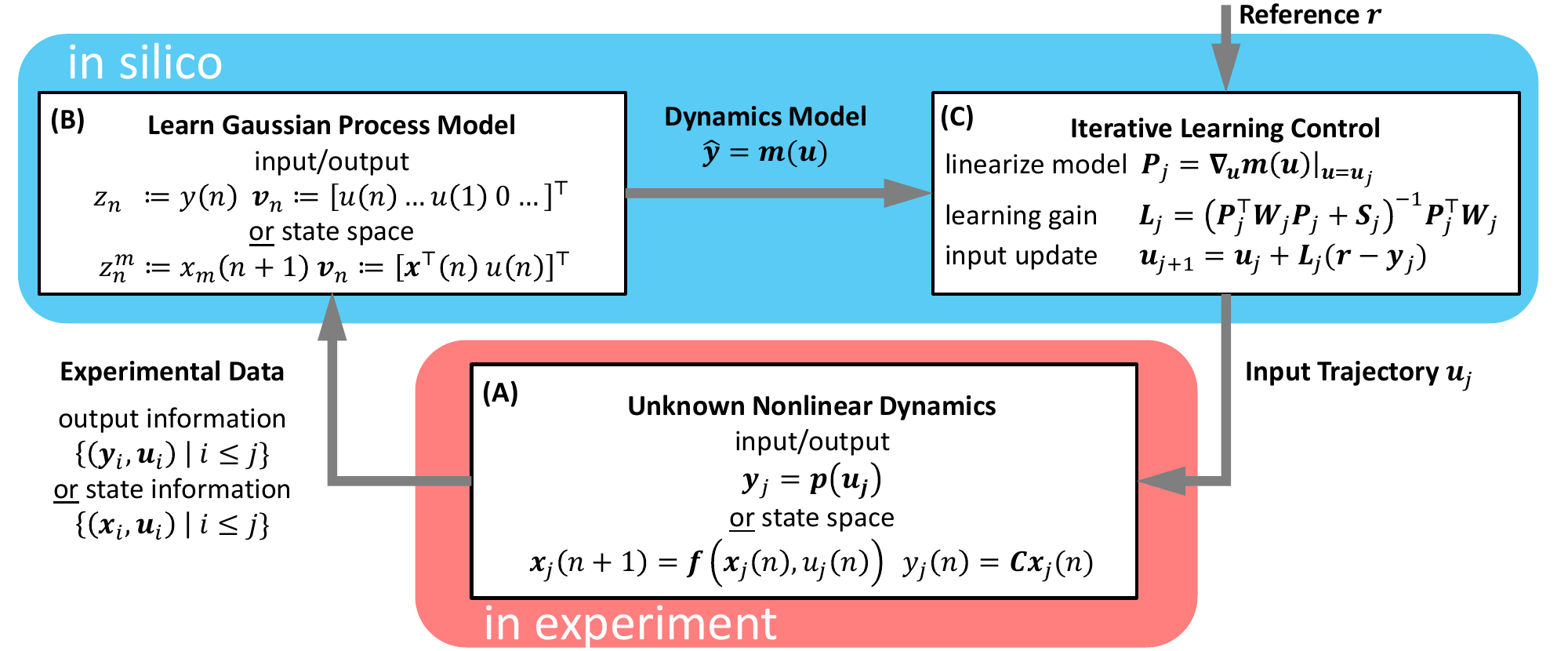}
    \caption{AI-MOLE is an iterative learning scheme, in which each iteration consists of three steps. (A) The current-trial input trajectory is applied to the unknown dynamics yielding either an output or state trajectory. 
    (B) The experimental data are then used to model the unknown dynamics using GPs. 
    (C) Finally, the GP model is utilized by an ILC update rule to compute the next-trial input trajectory.
    The iterations continue until the output trajectory converges sufficiently close to the reference trajectory.}
    \label{fig:method_overview}
\end{figure*}
We address the proposed problem with an iterative learning scheme called \emph{Autonomous Iterative Motion Learning} (AI-MOLE) that is autonomous in the sense that it automatically tunes its parameters even in the presence of unknown, nonlinear dynamics.
Each of the proposed method's iterations consists of three steps, see Figure \ref{fig:method_overview}.
First, a model of the plant dynamics is identified using the experimental data of previous trials, see Section \ref{sec:model_learning}.
To accommodate for the possibly nonlinear dynamics, a generic Gaussian process (GP) model is employed, which predicts the output trajectory for a given input trajectory.
Second, the next-trial input trajectory is determined by solving an optimal feedforward control problem based on the GP model, see Section \ref{sec:input_learning}.
Third, the updated input trajectory is applied to the plant and the resulting data is in turn used to refine the GP model.
Note that AI-MOLE only requires the data resulting from the single trial per iteration and that the updating of the GP model and learning of the input trajectory occur in \emph{parallel}.
To enable plug-and-play application, the proposed method framework autonomously determines necessary parameters, see Section \ref{sec:autonomous_parameters}.
\subsection{Model Learning}\label{sec:model_learning}
We propose a model, formally a function $\qm:\qRealNumbers^N\mapsto \qRealNumbers^N$, that predicts the plant's output trajectory $\qyest\in\qRealNumbers^N$ based on an input trajectory $\qu\in\qRealNumbers^N$, where the trial index is omitted to simplify notation.

As a model, we utilize GPs and thus we proceed by briefly reviewing their fundamentals.
Let $\qf(\qv):\qRealNumbers^D\mapsto \qRealNumbers$ denote the unknown target function that depends on the regression vector $\qv\in\qRealNumbers^D$.
Predictions are based on $K\in\qNaturalNumbers$ observations $z_k\in\qRealNumbers$ stemming from
\begin{equation}
    \forall k\in[1,K],\quad z_k = f(\qv_k) + \omega_k \quad \vert ~~ \omega_k\sim\mathcal{N}(0,\sigma^2_\omega)\,.
\end{equation}
The $K$ observation pairs $(z_k, \qv_k)$ are collected in the target vector $\qztraining\in\qRealNumbers^K$ and the design  matrix $\qVtraining\in\qRealNumbers^{D\times K}$, i.e.,
\begin{align}
    \qztraining &:= \begin{bmatrix}
    z\idx{1} & z\idx{2} & \dots & z_K
    \end{bmatrix}^\top\,,
    \\
    \label{eq:regression_matrix} \qVtraining &:= \begin{bmatrix}
    \qv\idx{1} & \qv\idx{2} & \dots & \qv_K
    \end{bmatrix}\,.
\end{align}
The kernel function of two  vectors $\qv\in\qRealNumbers^D$ and $\hat{\qv}\in\qRealNumbers^D$ is denoted by $k_{\qv\hat{\qv}}:\qRealNumbers^{(D,D)}\mapsto \qRealNumbers$. 
The kernel matrix of two  matrices $\qV\in\qRealNumbers^{D\times K}$, $\hat{\qV}\in\qRealNumbers^{D\times \hat{K}}$, which are assembled according to \eqref{eq:regression_matrix}, is denoted by $\qK_{\qV\hat{\qV}}\in\qRealNumbers^{K\times\hat{K}}$ and has entries $\left[\qK_{\qV\hat{\qV}}\right]_{ij} = k_{\qv_i\hat{\qv}_j}$.
        
Given $F\in\qNaturalNumbers$ test vectors assembled in the matrix $\qV\in\qRealNumbers^{D\times F}$, the predicted mean $\qpredictedmean\in\qRealNumbers^F$ and covariance $\qpredictedvar\in\qRealNumbers^{F\times F}$ are given by
\begin{align}\label{eq:predicted_mean}
    \qpredictedmean &= \qK_{\qV\qVtraining}\left[\qK_{\qVtraining\qVtraining}+\sigma^2_\omega\qI\right]^{-1}\qztraining
    \\
    \qpredictedvar &= \qK_{\qV\qV} - \qK_{\qV\qVtraining}\left[\qK_{\qVtraining\qVtraining}+\sigma^2_\omega\qI\right]^{-1}\qK_{\qVtraining\qV}\,.
\end{align}

In this work, we employ the GP framework to build a model $\qm$ that predicts the plant's output trajectory $\qy$ based on an input trajectory $\qu$.
For this purpose, we propose two different ways of employing the general GP framework to model the unknown dynamics.
Namely, we propose 1) an input/output model if only the plant's output variable is known and measured, referred to as the \emph{input/output(IO)-version} of AI-MOLE, and 2) a state-space model if the plant's state vector is known and measured, referred to as the \emph{input/state(IS)-version} of AI-MOLE.
Hereby, the model's characteristics are determined by the definition of the observation variable $z$, regression vector $\qv$, and kernel function $k$.

If only the plant's output variable is known and measured, we employ the GP framework to model the lifted dynamics \eqref{eq:lifted_dynamics}.
For this purpose, the observation variable is defined as the output variable, i.e.,
\begin{equation}\label{eq:io_model_observation}
    \forall n\in[1,N], \quad z_n := y(n)\,,
\end{equation}
and the regression vector consists of the current and all previous input samples, i.e.,
\begin{equation} \label{eq:regression_vector_io}
    \forall n\in[1,N], \quad \qv_n := \begin{bmatrix}
    u(n-1) & \dots u(1) & 0 & \dots & 0
    \end{bmatrix}^\top\,.
\end{equation}
 As kernel function, we employ a squared-exponential kernel (SEK) with a single length scale $l\in\qRealNumbers$, i.e.,
\begin{equation}
    k(\qv, \hat{\qv}) = \exp\left(-\frac{1}{2l^2}\left(\qv-\hat{\qv}\right)^\top\left(\qv-\hat{\qv}\right)\right)\,.
\end{equation}
To predict the output trajectory $\hat{\qy}$ for a given input trajectory $\qu$, we assemble the regression matrix $\qV$ according to \eqref{eq:regression_matrix} and \eqref{eq:regression_vector_io} and compute the predicted mean according to \eqref{eq:predicted_mean}, which is equal to the predicted output trajectory $\hat{\qy}$.

If the plant's state vector $\qx$ is known and measured, we employ the GP framework to model the state-space dynamics \eqref{eq:statespace_dynamics}.
For this purpose, we train $M$ GPs with each predicting the next sample of a respective state variable such that the $m^\mathrm{th}$ GP's observation variable is defined as
\begin{equation}
    \forall n\in[1,N],~ m\in[1,M],\quad z^m_n := \left[\qx(n+1)\right]_m\,,
\end{equation}
and the regression vector consists of the current state sample $\qx$ and input sample $u$, i.e.,
\begin{equation}\label{eq:regression_vector_is}
    \forall n\in[1,N], \quad \qv_n := \begin{bmatrix}
    \qx(n)^\top & u(n)
    \end{bmatrix}^\top\,.
\end{equation}
As kernel function, we employ a SEK with $M$ length scales $\forall m\in[1,M], l_m\in\qRealNumbers$, i.e.,
\begin{multline}
    k(\qv, \hat{\qv})=\exp \left(-\frac{1}{2} \left(\qv-\hat{\qv}\right)^\top \boldsymbol{\Lambda}^{-2}\left(\qv-\hat{\qv}\right)\right) \quad \vert \\ \boldsymbol{\Lambda} = \mathrm{diag}\left(l\idx{1}, l\idx{2}, \dots, l_M\right)\,.
\end{multline}
To predict the output trajectory $\qyest$ for a given input trajectory $\qu$, we use roll-out predictions to predict the progression of the state vector $\qx$ over samples and the output equation \eqref{eq:output_equation} to determine the estimated output trajectory $\qyest$.

For both models, the measurement variance $\sigma^2_\omega$ and length scales $l$ are determined using evidence maximization \cite{RasmussenW06}.
Because GP predictions become computationally expensive with increasing amounts of training data~\cite{RasmussenW06}, we limit the training data to the last $H\in\qNaturalNumbers$ trials.

Note that in this section only single-input/single-output dynamics were considered.
However, the proposed method can also be applied to multi-input/multi-output dynamics with $R\in\qNaturalNumbers$ input variables and $P\in\qNaturalNumbers$ output variables.
In case only the plant's output variables are known and measured, we employ $P$ separate GPs to predict each output variable according to (12) and the regression vectors contain the samples of all $R$ input variables according to (13).
In case the state vector is known and measured, the output matrix $\qC$ has $P$ rows to predict the output variables and the regression vector contains the samples of all $R$ input variables according to (16).

\subsection{Input Learning}\label{sec:input_learning}
After the GP model has been identified, it is used in an ILC update rule to determine the next-trial input trajectory that leads to a reduced next-trial error norm.
We employ a linear, trial-varying update law of the form 
\begin{equation} \label{eq:input_update_law}
    \forall j\in\qPositiveNaturalNumbers, \quad \qu_{j+1} = \qu_j + \qdeltau_j \quad \vert ~ \qdeltau_j=\qL_j\qe_j\,,
\end{equation}
where $\qL_j\in\qRealNumbers^{N\times N}$ is the trial-varying learning gain matrix.
To design the learning gain matrix, we first linearize the nonlinear GP model $\qm$ at the current input trajectory $\qu_j$, i.e,
\begin{equation}
    \forall j\in\qPositiveNaturalNumbers, \quad \qP_j := \qgrad_\qu \qm(\qu)\vert_{\qu=\qu_j}\,,
\end{equation}
and assume linearity, i.e.,
\begin{equation}
    \hat{\qy}_{j+1} = \qy_j + \qP_j \qdeltau_j\,.
\end{equation}
Given the linearization $\qP_j$, the learning gain matrix is designed via norm-optimal ILC, where the input update $\qdeltau_j$ is chosen to minimize the next-trial cost criterion
\begin{equation}
    J_j(\qdeltau_j) := \qe_{j+1}^\top \qW_j \qe_{j+1} + \qdeltau_j^\top \qS_j\qdeltau_j\,,
\end{equation}
where $\qW_j, \qS_j\in\qRealNumbers^{N\times N}$ are trial-varying, symmetric, positive-definite weighting matrices, such that
\begin{equation}
    \qdeltau_j = \underset{\qdeltau}{\mathrm{argmin}}~J_j(\qdeltau)\,.
\end{equation}
The optimization problem can be solved analytically leading to a linear update law of the form \eqref{eq:input_update_law} and the learning gain matrix
\begin{equation}\label{eq:learning_gain_design}
    \forall j\in\qPositiveNaturalNumbers,\quad \qL_j = \left(\qP_j^\top \qW_j\qP_j + \qS_j\right)^{-1}\qW_j \qP_j^\top\,.
\end{equation} 

\subsection{Autonomous Parameterization}\label{sec:autonomous_parameters}
To realize plug-and-play application, autonomous learning methods should not require manual tuning. In the following, we outline the procedure that enables AI-MOLE, in contrast to prior work, to autonomously determine necessary parameters.

First, we consider the choice of the initial input trajectory $\qu\idx{1}$ that is applied on the very first trial and generally is a robust parameter, i.e., that a large variety of choices of $\qu\idx{1}$ lead to successful learning.
To select $\qu\idx{1}$, we determine the largest significant frequency $f\idx{0}(\qr)$ of the reference trajectory $\qr$, and design a zero-phase low-pass filter $\qf\idx{LP}$ with cut-off frequency $f\idx{0}$.
The low-pass filter $\qf\idx{LP}$ is applied to a zero-mean normal distribution with covariance $\sigma^2\idx{I}\qI$, and the initial input trajectory is drawn from the resulting distribution, i.e.,
\begin{equation}\label{eq:initial_input}
    \qu\idx{1} \sim \qf\idx{LP}\left(\mathcal{N}(\qvec{0}, \sigma^2\idx{I}\qI)\right)\,.
\end{equation}
The input variance $\sigma^2\idx{I}$ is set to a small value such that the system is excited above the level of the measurement noise.
The value of the input variance $\sigma^2\idx{I}$ can be autonomously determined by a trivial procedure such as iteratively incrementing the input variance until the system excitation exceeds the measurement noise.

Once a trial is performed and the GP model is trained, the weights $\qW_j$ and $\qS_j$ are autonomously determined based on the GP model.
In case the input/output model is employed, the weighting matrices are chosen as
\begin{equation}\label{eq:io_weights}
    \forall j\in\qNaturalNumbers, \quad \qW^\mathrm{IO}_j=\qI, \quad \qS^\mathrm{IO}_j=\qTwoNorm{\qP_j}^2\qI\,,
\end{equation}
where $\qTwoNorm{\qP_j}$ is the induced Euclidean norm of the matrix $\qP_j$.
In case the state-space model is employed, the weighting matrices are chosen as
\begin{equation}\label{eq:ss_weights}
    \forall j\in\qNaturalNumbers, \quad \qW^\mathrm{IS}_j=\qI, \quad \qS^\mathrm{IS}_j=0.1\qTwoNorm{\qP_j}^2\qI\,.
\end{equation}

In order to reduce computational requirements, the number of trials that are used to train the GP model is set to $H=3$.

Note that the parameters $\qW$ and $H$ are set to constant one-size-fits-all values, i.e., AI-MOLE with the chosen values leads to good performance across a variety of real-world systems and tasks as will be explicitly shown later.
The parameter $\qS$ is \emph{autonomously} adjusted by AI-MOLE depending on the current trial's model, and, hence, the parameter depends on the current real-world system and task.
The combination of robust one-size-fits-all and self-tuning parameters makes AI-MOLE plug-and-play applicable because it does not require any \emph{manual} parameter tuning for a specific application.

\section{Experimental Performance Evaluation}
\begin{figure*}
    \centering
    \includegraphics[width=0.95\linewidth, trim={0cm 1cm 0cm 2cm}, clip]{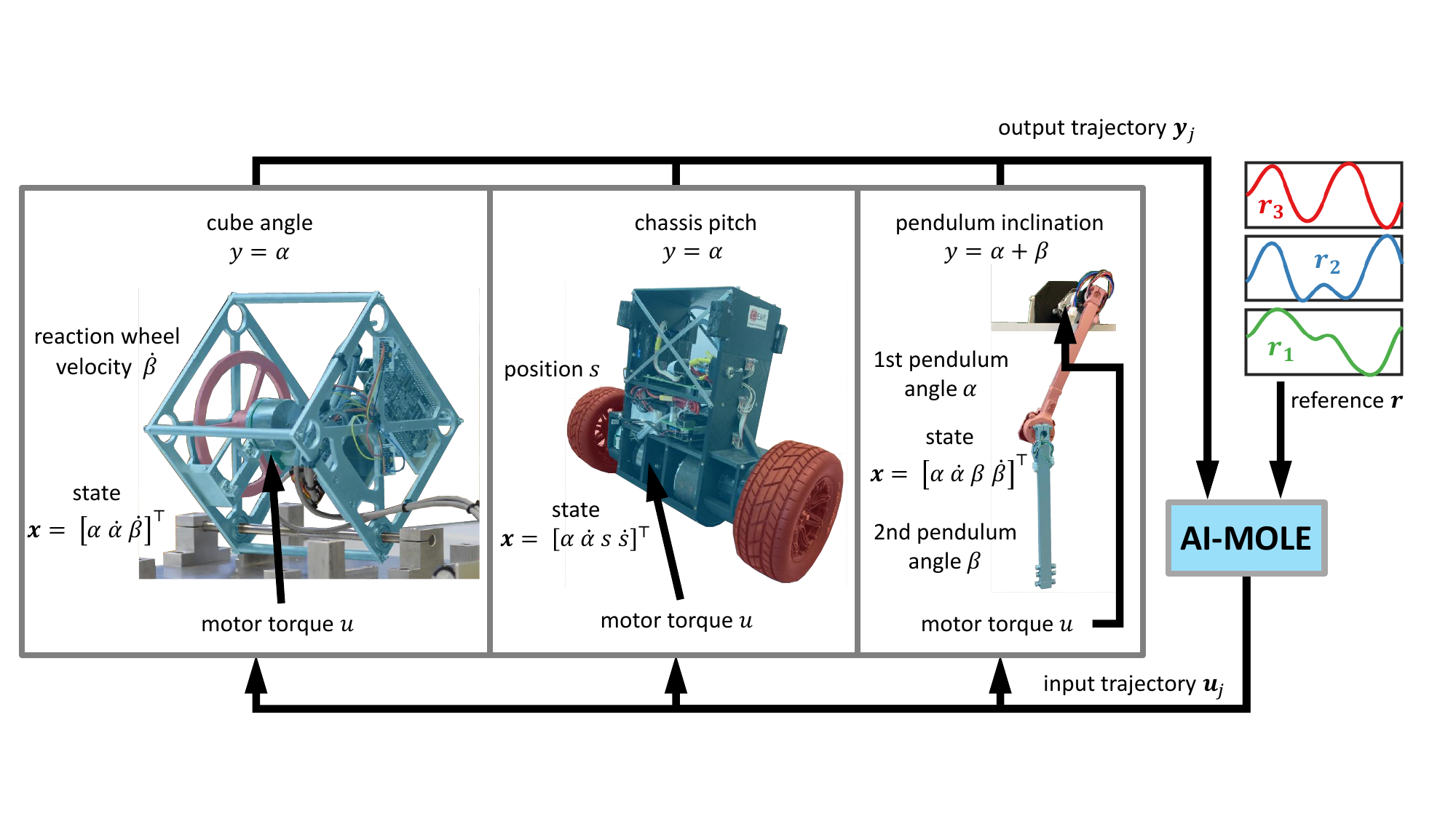}
    \caption{The experimental evaluation of AI-MOLE consists of three real-world testbeds, namely, the balancing cube (CUBE),  the two-wheeled inverted pendulum robot (TWIPR), and the double pendulum (PENDU). For each of the systems, AI-MOLE has to learn to track three different reference trajectories.}
    \label{fig:exp_prob}
\end{figure*}

In this section, both versions of AI-MOLE are validated on three different real-world systems and three different reference tracking tasks per system, see Figure \ref{fig:exp_prob}.
First, the real-world testbeds and reference tracking tasks are described.
Second, the results of the IO-version of AI-MOLE  are presented to demonstrate the method's capability of solving reference tracking tasks in a plug-and-play fashion using only input/output information.
Third, the IS-version of AI-MOLE is demonstrated to yield even better learning performance if state information is available. 
\subsection{Testbeds}
We consider a total of three different systems, namely, the pendulum-cube (CUBE), the two-wheeled inverted pendulum (TWIPR), and the double pendulum (PENDU), see Figure \ref{fig:exp_prob}.
For all three of the systems complex, nonlinear models of the nominal dynamics have been derived \cite{dynamic_twipr, 6385896, MAITI2016408}, but note that the actual real-world systems are affected by additional dynamic phenomena such as stick-slip friction, backlash of gearings, and minor elasticity of 3D-printed materials.
 
The first system, CUBE, consists of the cube-shaped pendulum body and the reaction wheel that is driven by a DC motor which is mounted to the cube.
The cube's angle is denoted by $\alpha\in\qRealNumbers$ and serves as the output variable, the reaction wheel's angle is denoted by $\beta\in\qRealNumbers$, and the torque produced by the motor is denoted by $u\in\qRealNumbers$ and serves as the input variable.
The system's state vector consists of the cube's angle, the cube's angular velocity, and the angular velocity of the reaction-wheel, i.e.,
\begin{equation}
    \forall j,~n\in\qPositiveNaturalNumbers, \quad \qx_j(n):=\begin{bmatrix}
    \alpha_j(n) & \dot{\alpha}_j(n) & \dot{\beta}_j(n)
    \end{bmatrix}^\top\,.
\end{equation}
The system's output matrix is given by
\begin{equation}
    \qC := \begin{bmatrix} 1 & 0 & 0 \end{bmatrix}\,.
\end{equation}

The second system, TWIPR, consists of a pendulum body housing main electronics including a microcomputer, inertial measurement units, motors, and battery. 
Wheels are mounted onto the motors such that the robot can drive while balancing its chassis. 
To stabilize the robot in its upright equilibrium, a linear state feedback controller is employed \cite{10.3389/frobt.2022.793512}. 
The pitch angle of the robot is denoted by $\Theta\in\qRealNumbers$ and the position of the robot is denoted by $s\in\qRealNumbers$, the pitch angle serves as output variable, i.e., $y:=\Theta$, and the motor torque $u\in\qRealNumbers$ that drives the wheels serves as the input variable. 
The system's state vector consists of the robot's pitch angle, pitch velocity, position, and linear velocity, i.e.,
\begin{equation}
    \forall j, n\in\qPositiveNaturalNumbers, \quad \qx_j(n):=\begin{bmatrix}
    \Theta_j(n) & \dot{\Theta}_j(n) & s_j(n) & \dot{s}_j(n)
    \end{bmatrix}^\top\,.
\end{equation}
The system's output matrix is given by
\begin{equation}
    \qC := \begin{bmatrix} 1 & 0 &0 & 0 \end{bmatrix}\,.
\end{equation}

The third system, PENDU, is a double pendulum, whose first link is driven by a DC motor.
The angle of the first pendulum is denoted by $\alpha\in\qRealNumbers$, and the relative angle of the second pendulum with respect to the first pendulum is denoted by $\beta\in\qRealNumbers$.
The absolute angle of the second pendulum serves as the output variable, i.e., $y:=\alpha+\beta$, and the torque produced by the motor is denoted by $u\in\qRealNumbers$ and serves as input variable.
The system's state vector consists of the first and second  pendulum's angle and angular velocity, i.e,
\begin{equation}
    \forall j,~ n\in\qPositiveNaturalNumbers, \quad \qx_j(n):=\begin{bmatrix}
    \alpha_j(n) & \dot{\alpha}_j(n) & \beta_j(n) & \dot{\beta}_j(n)
    \end{bmatrix}^\top\,.
\end{equation}
The system's output matrix is given by
\begin{equation}
    \qC := \begin{bmatrix} 1 & 0 & 1 & 0 \end{bmatrix}\,.
\end{equation}

All of the three systems are sampled at a frequency of 50 Hz, and for each of the three systems, AI-MOLE is evaluated using three different reference trajectories that greatly differ in frequency and amplitude. To ensure that each of the references is realizable, the references are generated by applying an input trajectory to the respective system and using the resulting output trajectory as the reference. The input trajectories are generated by drawing a random trajectory from a zero-mean normal distribution and applying a zero-phase low-pass filter to the random trajectory.

In contrast to the assumption made in the problem formulation, the real-world testbeds naturally do not have trial-invariant dynamics, and, hence, the tracking error cannot be expected to converge to zero.
Instead, tracking precision can only be expected to reach the system's repeatability. 
To judge the learning method's performance independently of the system's repeatability and the reference's amplitude, we use the relative error  
\begin{equation}
\epsilon_j := \left\{ \begin{matrix}\frac{\qTwoNorm{\qe_j}}{\qTwoNorm{\qr}}- e_\mathrm{R} & \mathrm{if} & \frac{\qTwoNorm{\qe_j}}{\qTwoNorm{\qr}} -e_\mathrm{R}>0
    \\
    0 & \mathrm{else} & ~ \end{matrix}\right.\,,
\end{equation}
which is limited to a minimum value of zero and where $\qreperror\in\qRealNumbers$ is the system's \emph{maximum repetitive error}. 
For each of the three experimental testbeds, the maximum repetitive error is determined via the following procedure. 
Recall that the input trajectories that lead to the output perfectly tracking the respective reference trajectory are known because the references were determined by applying known, randomly generated input trajectories to the respective system.
Now, for each of the three references, the corresponding input trajectory is applied to the system ten times and the output trajectories $\qy_i, ~ i\in[1,10]$, are recorded.
For each of the ten output trajectories, the relative deviation from the references is recorded, and the maximum of the respective error norms is considered the maximum repetitive error, i.e., 
\begin{equation}
    \qreperror := \underset{i\in[1,10]}{\mathrm{max}}~  \frac{\qTwoNorm{\qr - \qy_i}}{\qTwoNorm{\qr}}\,.
\end{equation}
For CUBE the repetitive error equals $5\%$, for TWIPR $12\%$, and for PENDU $10\%$.

Further note that two of the systems, namely CUBE and PENDU, are open-loop and do not employ an underlying feedback controller.
In contrast, TWIPR is a closed-loop system with an underlying feedback controller.
Hence, the experimental results also demonstrate that AI-MOLE can be applied to systems independent of whether they are feedback-controlled or not.
\subsection{Performance Without State Knowledge}
\begin{figure}
    \centering
    \includegraphics[width=\columnwidth, trim={0cm 0cm 0cm 0cm},clip]{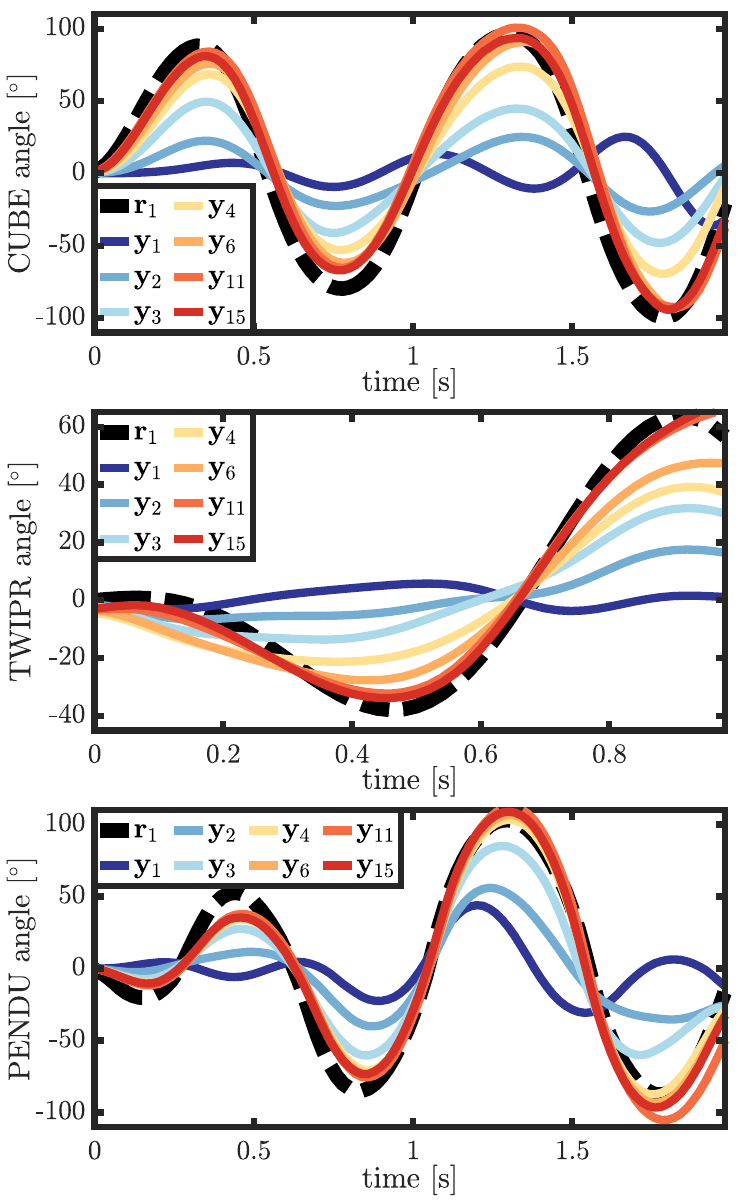}
    \caption{The output trajectories for learning the respective first reference for CUBE, TWIPR, and PENDU show that for each task the initial output trajectory is close to zero and significantly deviates from the reference. From there onwards, the output trajectories quickly converge to the reference independent of the system, trial length, and the references' amplitudes and frequencies.}
    \label{fig:io_tracking_performance}
\end{figure}

\begin{figure}
    \centering
    \includegraphics[width=\columnwidth, trim={0cm 0cm 0cm 0cm},clip]{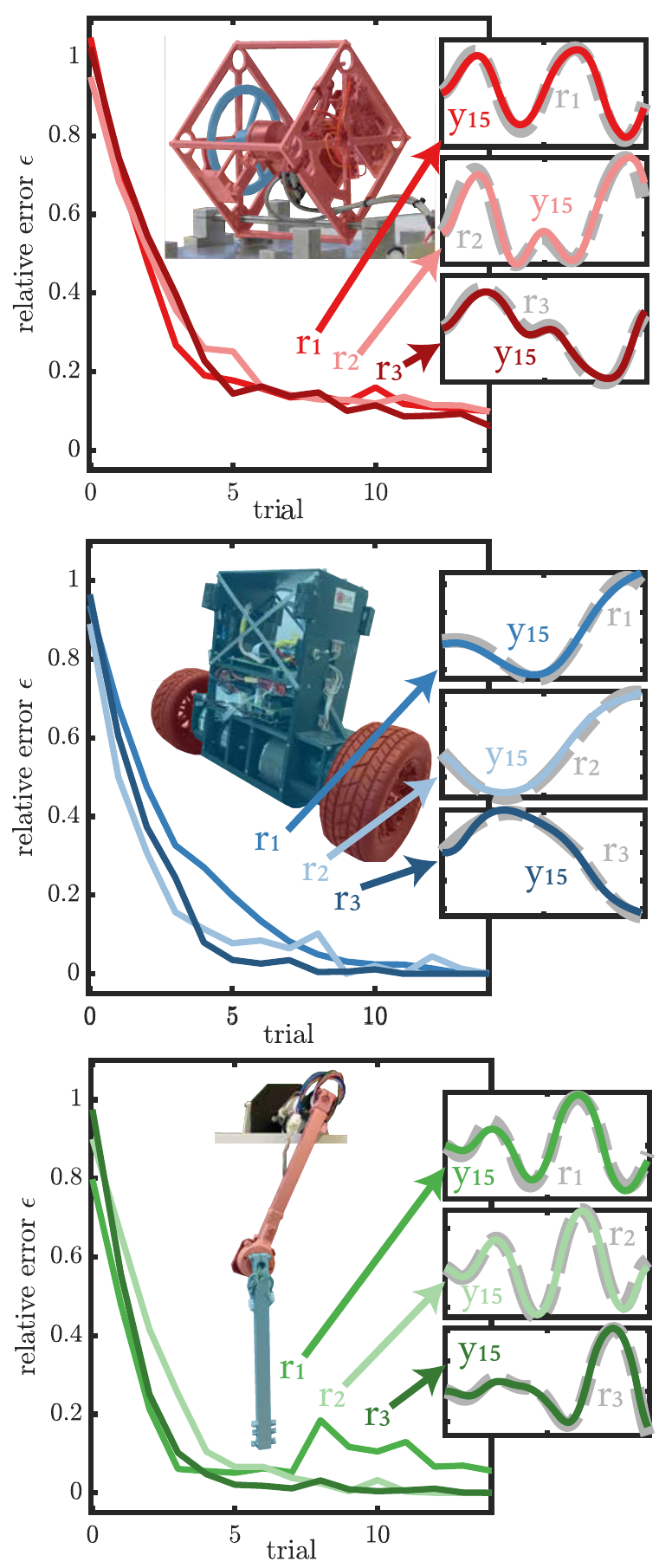}
    \caption{Experimental evaluation of AI-MOLE (IO-version) on three testbeds and, for each testbed, three respective reference trajectories.
    Despite varying dynamics and references, and without any manual tuning, satisfying tracking performance is achieved within 5--10 trials in all nine tasks. The results demonstrate that AI-MOLE is capable of autonomously solving reference tracking tasks for a variety of different systems under real-world settings in a truly plug-and-play fashion.}
    \label{fig:fir_does_it_all}
\end{figure}
In this section, the IO-version of AI-MOLE is evaluated on three experimental testbeds and, for each testbed, on three reference trajectories.

We start by outlining the general procedure of how AI-MOLE is applied.
Consider the CUBE testbed and the first reference trajectory.
First, AI-MOLE autonomously determines and applies the initial input trajectory $\qu\idx{1}$ according to \eqref{eq:initial_input}, which yields the output trajectory $\qy\idx{1}$. 
Next, AI-MOLE uses the observed data to determine the first GP-model of the unknown, nonlinear dynamics, which is in turn used to autonomously determine the weights and learning parameters \eqref{eq:io_weights}, \eqref{eq:learning_gain_design}. 
The latter is used in the update law \eqref{eq:input_update_law} to determine the next-trial input trajectory $\qu\idx{2}$.
From here onwards, AI-MOLE repeats the learning loop for a total of 15 trials.

The same procedure is repeated for the other systems and reference trajectories. 
Note especially and in stark contrast to almost all previous work that, here, the proposed method AI-MOLE is neither altered nor adjusted and that one and the same method - without any manual tuning - is applied to three completely different application systems.

The output trajectories depicted in Figure \ref{fig:io_tracking_performance} show that each of the initial output trajectories is close to zero and greatly differs from the respective reference trajectory.
However, the relative errors drop rapidly over the first few trials for all three real-world systems and all nine references, see Figure~\ref{fig:fir_does_it_all}.
AI-MOLE only requires 2--3 trials to reduce the initial error by 50\% and roughly five trials to reduce the initial error by 80\%. 
After roughly 10 trials, the relative errors for TWIPR and PENDU converge to the respective system's repetitive error, i.e., the best performance possible is achieved. 
Only for CUBE, does a minor deviation between the final relative errors and the system's repetitive error remains. 
However, comparing the output trajectories to their respective reference trajectories after 15 trials shows that not only for TWIPR and PENDU but also for CUBE, AI-MOLE achieves highly precise tracking.

In summary, the results demonstrate that AI-MOLE is capable of rapidly learning to solve reference tracking tasks in real-world systems while only requiring input/output information. 
Because the method was not only applied to a single benchmark system and task but to a total of nine references and three real-world systems, we conclude that AI-MOLE is robust and reliable and the results are replicable.
Despite being applied to different references and systems, AI-MOLE autonomously reconfigured necessary learning parameters such that learning happened in a truly plug-and-play fashion.

Note that AI-MOLE does not provide formal conditions to theoretically guarantee stability, and, hence, may be limited w.r.t. safety-critical applications.
Nonetheless, in the experiments, AI-MOLE has achieved stable learning and convergence on three different systems and a total of nine different tasks, which \emph{empirically indicates} the method's stability.

\subsection{Performance With State Knowledge}
\begin{figure}
    \centering
    \includegraphics[width=\columnwidth]{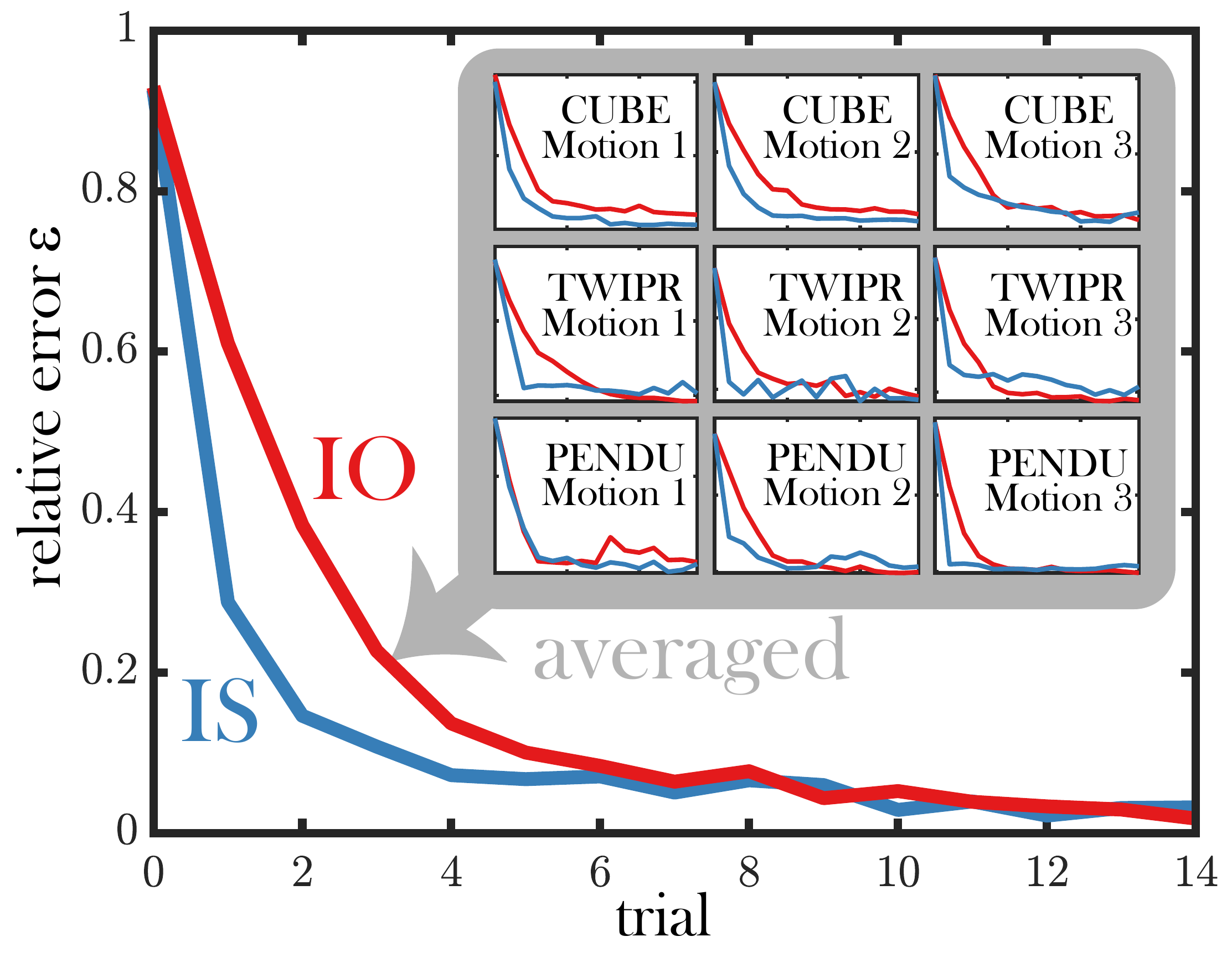}
    \caption{Both, the IO- and IS-version of AI-MOLE converge to low errors. However, the IS-version utilizes the additional state information to accelerate the speed of learning.}
    \label{fig:ss_accelerates_learning}
\end{figure}
In this section, we show that AI-MOLE can utilize state information if available, i.e., the entire state vector is measured, to accelerate the speed of learning.
For this purpose, the IS-version of AI-MOLE is applied to the three real-world testbeds and, for each testbed, three respective reference trajectories.
Note that once again one and the same method is applied to three different systems without any manual adjustment or tuning.

The resulting progressions of the relative errors are depicted in Figure \ref{fig:ss_accelerates_learning} and show that the IS-version of AI-MOLE is also capable of rapidly learning to track the nine reference trajectories in a plug-and-play fashion.
Furthermore, the comparison of the average learning progression of the versions shows that the IS-version's error declines more rapidly over the first trials.
On average, the IS-version of AI-MOLE decreases the error by more than 80\% after the first two trials and by more than 90\% after the first four trials.

We, hence, conclude that the IS-version of AI-MOLE is capable of not only learning to solve reference tracking tasks in a truly plug-and-play fashion but also accelerating learning by exploiting additional state information.

\section{Discussion and Conclusion}
In this work, we have proposed a learning method, called AI-MOLE, to solve reference tracking tasks in systems with unknown, nonlinear dynamics.
Unlike existing, state-of-the-art methods, AI-MOLE works in a truly plug-and-play fashion and does neither require task-specific prior knowledge nor manual parameter tuning.
On each iteration, AI-MOLE applies a feedforward input trajectory to the system, approximates the unknown dynamics with a Gaussian Process model, and uses the latter to update the input trajectory via an ILC update rule.
AI-MOLE only requires input/output information of a system, but can also exploit additional state information, if available, to accelerate the speed of learning.
Furthermore, AI-MOLE self-reliantly determines the necessary parameters to supersede manual parameter tuning.

In contrast to state-of-the-art learning methods, which either require manual parameter tuning, are restricted to simulated environments, or only have been validated on a single experimental testbed, AI-MOLE has achieved the unprecedented result of solving a total of nine different reference tracking tasks on three different real-world systems without requiring any manual parameter tuning and while only requiring 5-10 trials per task.
While other state-of-the-art methods may achieve comparable learning and tracking performance, AI-MOLE improves the state of the art by requiring no a priori model information, no system- or task-specific knowledge, and no manual parameter tuning.
Hence, AI-MOLE sets a milestone with respect to the autonomy of learning methods and the validation of said autonomy. 
We believe that these results make a significant contribution to advancing learning control methods such that they enable true autonomy in robots.

Despite these achievements, the proposed method is subject to the following limitations. 
First, AI-MOLE has been validated on single-input/single-output systems, with validation on real-world multi-input multi-output systems being part of our ongoing research.
Second, no formal conditions for convergence and stability of the learning method have been provided. Subsequent research will focus on theoretically investigating these properties.

\bibliographystyle{ieeetran}
\bibliography{aimole_bib}
\end{document}